\documentclass[10pt,twocolumn,letterpaper]{article}

\usepackage{iccv}
\usepackage{times}
\usepackage{epsfig}
\usepackage{graphicx}
\usepackage{amsmath}
\usepackage{amssymb}
\usepackage{color, colortbl}
\usepackage[accsupp]{axessibility}

\usepackage[pagebackref=true,breaklinks=true,letterpaper=true,colorlinks,bookmarks=false]{hyperref}

\usepackage{fmsymbols}
\usepackage[ruled,vlined,linesnumbered,resetcount]{algorithm2e}
\usepackage[capitalize]{cleveref}
\crefname{section}{Sect.}{Sects.}

\definecolor{Blue}{rgb}{0.88,1,1}
\newcolumntype{g}{>{\columncolor{Blue}}c}

\iccvfinalcopy 

\def\iccvPaperID{3587} 

\ificcvfinal\fi

\begin{document}

\title{Segmentation of Tubular Structures\\ Using Iterative Training with Tailored Samples}

\author{Wei Liao\\
  Independent Researcher\\
  {\tt\small liaowei.post@gmail.com}
}

\maketitle
\ificcvfinal\thispagestyle{empty}\fi

\begin{abstract}
  We propose a minimal path method to simultaneously compute segmentation masks and extract centerlines of tubular structures with line-topology.
  Minimal path methods are commonly used for the segmentation of tubular structures in a wide variety of applications.
  Recent methods use features extracted by CNNs, and often outperform methods using hand-tuned features.
  However, for CNN-based methods, the samples used for training may be generated inappropriately, so that they can be very different from samples encountered during inference.
  We approach this discrepancy by introducing a novel iterative training scheme, which enables generating better training samples specifically tailored for the minimal path methods without changing existing annotations.
  In our method, segmentation masks and centerlines are not determined after one another by post-processing, but obtained using the same steps.
  Our method requires only very few annotated training images.
  Comparison with seven previous approaches on three public datasets, including satellite images and medical images, shows that our method achieves state-of-the-art results both for segmentation masks and centerlines.
\end{abstract}

\section{Introduction}
\label{sec:intro}

\begin{figure}[!tb]
  \centering
  \begin{tabular}{cc}
    \includegraphics[width=3.6cm]{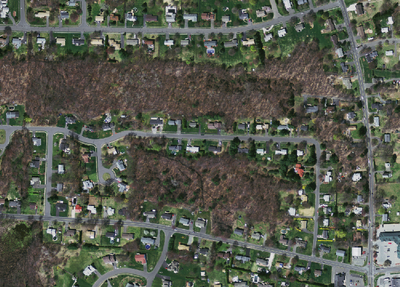}&
    \includegraphics[width=3.6cm]{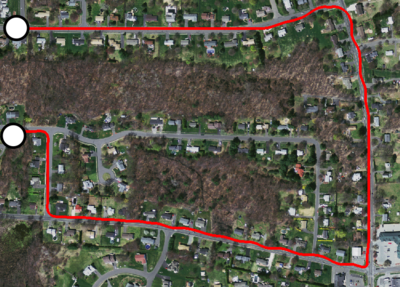}\\
    \includegraphics[height=3.6cm, angle=90]{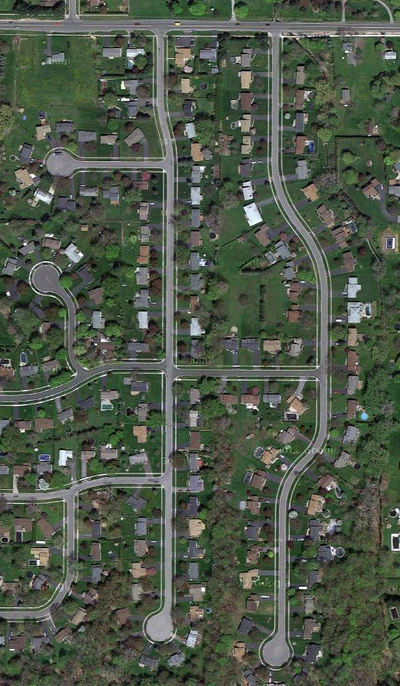}&
    \includegraphics[height=3.6cm, angle=90]{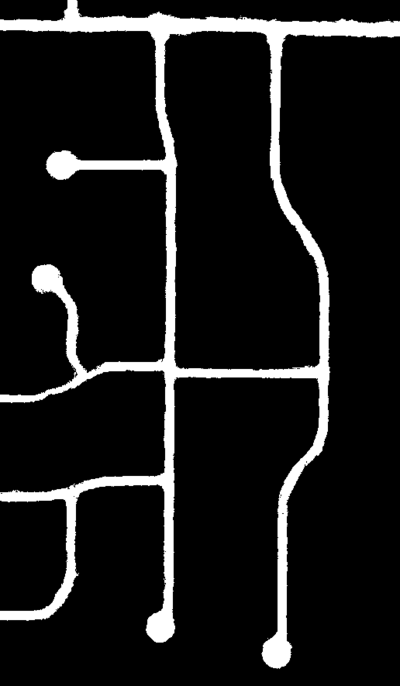}\\
    \includegraphics[width=3.6cm]{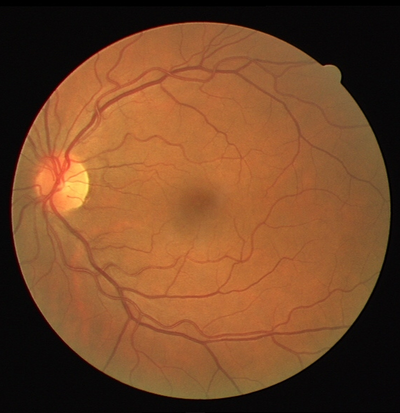}&
    \includegraphics[width=3.6cm]{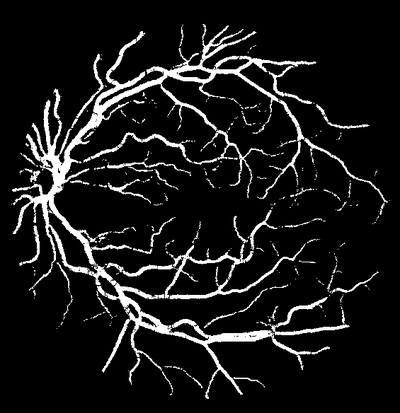}\\
    (a) Input image &  (b) Our results
  \end{tabular}
  \caption{Examples of application areas of tubular structure segmentation.
    %
    First row: Finding path between two points in a satellite image (navigation).
    Second row: Segmentation of road network in a satellite image.
    Third row: Segmentation of the vessel tree in a image of retinal vessels.
  }
  \label{pic:application_area}
\end{figure}

\begin{figure}[!tb]
  \centering
    \includegraphics[width=8cm]{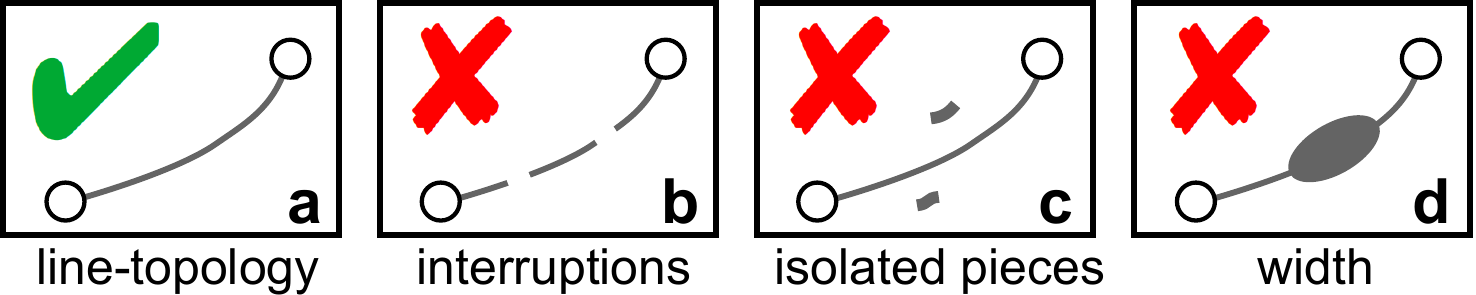}\\
  \caption{Illustration of line-topology and non-line-topology.
  }
  \label{pic:topology}
  \vspace*{-15pt}
\end{figure}

Segmentation of tubular structures is an essential task in many application areas of computer vision, including navigation, delineation of roads in satellite images, and analysis of blood vessels in medical images (see \cref{pic:application_area} for examples).
Beyond segmentation masks, such applications often also require centerlines with line-topology, i.e., the centerlines should not contain interruptions, isolated pieces, or width (\cref{pic:topology}).
Centerlines with line-topology are usually represented as sequences of coordinates.

While non-tubular object such as cars often have relatively high variance of appearance (e.g., color or texture) but low variance of shapes, tubular structures typically have relatively low variance of appearance but high variance of shapes.
For example, blood vessels in medical images of the same imaging modality typically have similar intensity distributions, but they may contain very irregular shapes caused by pathology.
General segmentation methods, such as U-Net \cite{Ronneberger_MICCAI_2015_U-Net}, are widely used also to segment tubular structures, but these methods do not take the special properties of tubular structures into account, and cannot satisfy certain topological constraints, such as connectivity of two points.
To obtain centerlines with line-topology, post-processing is usually required.
Also, the amount of annotations needed to train such methods is usually quite high.

In contrast, using minimal path methods, such as Dijkstra's algorithm \cite{Dijkstra59:NM} or the fast marching method \cite{Cohen_IJCV_1997,Sethian1999}, points on the centerlines  are obtained explicitly, and line-topology is guaranteed naturally.
However, to obtain segmentation masks of the tubular structures, often an additional post-processing step is necessary \cite{Freiman2009}.
Alternatively, a new dimension can be introduced into the parameter space of the minimal path method to represent the radius \cite{Li2007,Pechaud2009,Chen_TIP_2019}, but this leads to  significantly higher computational cost.
In the recent method of \pathcnn \cite{Liao_CVPR_2022_Progressive}, a CNN is integrated into Dijkstra's algorithm to obtain centerlines and masks without post-processing.
However, the masks generated using this method is relatively imprecise, since training is only performed using samples along user-specified centerlines, i.e., samples which are inside the ground truth masks of tubular structures but do not lie exactly on centerlines are not used.
Furthermore, the samples used to train this CNN are often different from samples which are actually encountered during the inference.

We propose a method to determine segmentation masks and centerlines with line-topology for tubular structures in a unified way.
Our method uses a similar framework as \cite{Liao_CVPR_2022_Progressive}, i.e., we integrate a CNN classifier into the minimal path method.
However, we use a novel technique for creating samples out of existing annotations (see \cref{pic:inference_scheme} and \ref{pic:training_scheme} for an overview of differences).
Instead of using the annotation (i.e., ground truth mask) directly and training the model only once, we introduce an \emph{iterative training scheme}.
In each iteration, we apply minimal path method with the CNN classifier trained in the previous iteration to generate new samples out of the annotations.
These samples are used to re-train the CNN classifier in the current iteration.
The ground truth masks do \emph{not} change, but by iteratively re-training and re-sampling, we are able to extract \emph{different} samples which are much more realistic and better \emph{tailored} for the minimal path method.
To our knowledge, this is the first approach to apply iterative training to minimal path method, and to use minimal path method itself to improve samples during the training phase.
In this way, we better exploit the properties of the minimal path method, and achieve more precise masks and centerlines on three datasets.
Also, our method needs only very few annotated images, making it especially effective for medical image analysis, for which the amounts of annotations are often limited.

\section{Related Work}
\paragraph{Topology enhancement}
Local features in the neighborhood of a pixel can be used to estimate the \emph{tubularity measure}, i.e., probability that this pixel belongs to a tubular structure.
Several widely used methods employ differential measures based on Hessian matrix \cite{Frangi1998,Sato1998} or flux \cite{Tueretken_ICCV_2013}.
While most of these methods rely on hand-tuned parameters, there also exist learning-based methods, such as \cite{Sironi_PAMI_2015b}.
These methods are usually used by other methods as feature extraction step to achieve segmentation masks or centerlines.
More recently methods \cite{Hu_NIPS_2019_TopologyPreservingDeepImageSegmentation,Shit_CVPR_2021_clDice} use deep learning to enhance tubular structures, but post-processing is still needed to achieve line-topology of the result.

\paragraph{Approaches based on minimal path}
Minimal path methods are commonly used to extract centerlines of tubular structures with line-topology.
These methods often suffer from the short cut problem, i.e., the centerlines may leak out the tubular structures.
This can especially happen when the tubular structure has high curvature, or two tubular structures are very close to each other, or the tubularity measure cannot differentiate between foreground and background.
The short cut problem has been approached using a variety of methods \cite{Benmansour2009b,Benmansour_IJCV_2011,Kaul-Yezzi-Tsai_PAMI_2012,Ulen_PAMI_2015,Li2007,Pechaud2009,Chen_TIP_2019}.
The recent method \cite{Liao_CVPR_2022_Progressive} integrates a CNN classifier into the minimal path method to avoid several types of short cut problems.
However, the classifier in \cite{Liao_CVPR_2022_Progressive} is only trained once, and minimal path method is only used during inference.
In contrast, our method is trained iteratively, and minimal path method is used both during training and inference.
Furthermore, in \cite{Liao_CVPR_2022_Progressive} a hand-tuned tubularity feature \cite{Frangi1998} is used at initialization, but in later steps other learning-based features are used.
In contrast, we use learning-based features throughout all steps, thus it is not necessary to tune parameters manually for \cite{Frangi1998}.

\paragraph{Training scheme}
Instead of using annotations directly, \cite{Mosinska_PAMI_2020} uses more realistic samples provided by a segmentation network.
We use a similar idea, but in our approach, the training process is iterative, i.e., the classifier is trained using samples which are generated in the previous iteration \emph{using the classifier itself} in conjunction with the minimal path method.
The classifier in \cite{Mosinska_PAMI_2020} is only used to classify samples created by \emph{another} segmentation network, while our classifier is an \emph{integral part} of the segmentation method itself.
Furthermore, \cite{Mosinska_PAMI_2020} relies on other tubularity features, which require large amounts of training data. In contrast, our method does not depend on other features and yields high-quality results even for small amounts of training data.
Finally, \cite{Mosinska_PAMI_2020} does not achieve segmentation masks.

\paragraph{Road segmentation}
There are recent methods for the segmentation of road networks \cite{Mosinska_PAMI_2020,Bastani_CVPR_2018,Batra_CVPR_2019_RoadJointLearning}, and some of them use minimal path method as a component \cite{Mosinska_PAMI_2020}.
However, our focus is on the methodology itself, and our method can also be applied to other images, such as medical images.

\section{Minimal Path Method with Integrated CNN}
Minimal path methods can be naturally used to segment centerlines of tubular structures.
These methods are efficient since they are usually based on Dijkstra's algorithm \cite{Dijkstra59:NM}, which is a greedy algorithm, or its continuous counterpart, the fast marching method \cite{Cohen_IJCV_1997,Sethian1999}.
Our approach is based on Dijkstra's algorithm, but the same idea also applies to the fast marching method.

\subsection{Centerline Extraction}

\begin{figure}[!t]
  \centering
  \begin{tabular}{cc}
    \includegraphics[width=3.6cm]{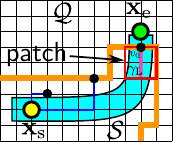}&
    \includegraphics[width=3.6cm]{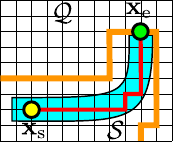}\\
    (a) & (b)
  \end{tabular}
  \caption{
  	Segmentation of a tubular structure using Dijkstra's algorithm.
    (a) An intermediate step.
    (b) Path extraction: The set \dijS has reached the end point \fmendpoint, and the minimal path can be extracted using the predecessor function \dijPrev.
  }
  \label{pic:dijkstra}
\end{figure}

Suppose a graph $\dijGraph = (\dijVertex, \dijEdge)$ is induced by image \image, where \dijVertex and \dijEdge are the sets of \emph{vertices} and \emph{edges}, respectively.
Each vertex $v$ corresponds to a pixel in \image, so that \dijGraph can be imagined as a regular grid over the image.
The example in \cref{pic:dijkstra} shows such an image grid.
Neighboring pairs of vertices $u$ and $v$ are connected by an edge $e_{u,v}$, which has a weight $\dijW[u,v]$ depending on the underlying image feature.
For the tubular structure to be segmented, such as the cyan object in \cref{pic:dijkstra}, we assume that the start point \fmstartpoint and end point \fmendpoint are given.
The requirement for given start and end points is not a severe limitation, since in many applications these points can be determined using other methods automatically, such as the method in \cite{Mosinska_PAMI_2020}.
Finding the centerline of this tubular structure is equivalent to determining a path \mypath, i.e.,  a sequence of vertices $\{v_0, v_1, \dots, v_{|\mypath|}\}$, from \fmstartpoint to \fmendpoint in such a way that the total weight $\sum_{i=1}^{|\mypath|} \dijW[e_{v_{i-1}, v_i}]$ of \mypath is minimized among all possible paths between \fmstartpoint and \fmendpoint.
This path with minimum total weight is referred to as the \emph{minimal path}.

The minimal path is detected by computing the minimum total weight $d(u)$ of paths between \fmstartpoint and any other vertex $u$ in the graph.
In \cref{pic:dijkstra}, an orange curve partitions $\dijVertex$ into two regions: In the region $\dijS$, which contains \fmstartpoint, $d(u)$ is already determined for all vertices, while in the region $\dijQ$, $d(u)$ is not yet determined.
For every vertex $u$ inside $\dijS$, its \emph{predecessor} is recorded using the function $\dijPrev(u)$.
By repeatedly looking up the predecessor, one can always back-trace to \fmstartpoint.
Paths to \fmstartpoint are illustrated as blue lines exemplarily for several vertices (black dots) in \cref{pic:dijkstra}a.
In each iteration of the minimal path method, one vertex is added to $\dijS$, so that $\dijS$ always grows towards \fmendpoint, while \dijQ always shrinks.
Once \dijS reaches \fmendpoint, the minimal path can be found by tracing from \fmendpoint back to \fmstartpoint using \dijPrev (red line in \cref{pic:dijkstra}b).
For details, we refer to standard texts such as \cite{Cormen2009,Kleinberg2005}.

\subsection{Segmentation Mask with \pathcnn}
\label{sec:pathcnn}

\begin{figure}[!t]
  \centering
  \begin{tabular}{cc}
    \includegraphics[width=3.8cm]{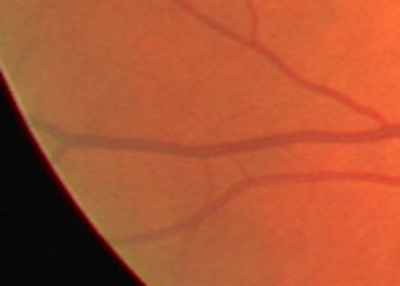}&
    \includegraphics[width=3.8cm]{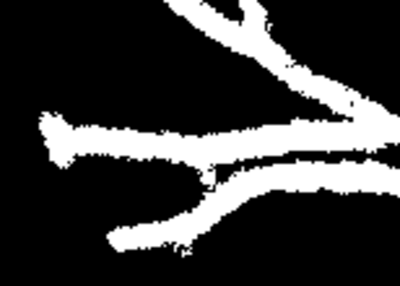}\\
    (a) Input image & (b) Segmentation mask
  \end{tabular}
  \caption{Result of \pathcnn on an image of retinal vessels from \datadrive.
    (a) Input image.
    (b) The segmentation mask is too large.
  }
  \label{pic:pathcnn_boundary_problem}
\end{figure}

In the \pathcnn method \cite{Liao_CVPR_2022_Progressive}, an additional step is introduced into Dijkstra's algorithm.
Each time when a vertex $v_0$ is added to \dijS, a short path \mypathlocal (magenta line in \cref{pic:dijkstra}a) is back-traced using \dijPrev.
\mypathlocal has only a fixed length and usually does not end in \fmstartpoint.
Then an image patch \mypatch (red box) is cropped around \mypathlocal.
If \mypathlocal is curved, then \mypatch is also curved.
\mypatch is referred to as \emph{sample}.
Then \mypatch is rectified, i.e., warped into a rectangular patch \mypatchwarp so that \mypathlocal becomes a \emph{straight vertical line} in the middle of \mypatchwarp.
The patch in \cref{pic:dijkstra}a is the same as its rectified version.
After that, \mypatchwarp is classified by a CNN classifier \myclassifier.
Further steps depend on the classification result, and more details are given in \cref{sec:inference} below.
Since \pathcnn not only computes \dijPrev but also classifies each pixel in the image as foreground or background, consequently its result contains not only \emph{centerlines} of tubular structures, but also a binary \emph{segmentation mask} of them.
However, since \pathcnn is trained directly using samples along ground truth centerlines, its classification performance for pixels off the centerlines is not strong, and often the boundary of the tubular structures is not delineated exactly.
An example is shown in \cref{pic:pathcnn_boundary_problem}, in which the mask for retinal vessels generated by \pathcnn is too large.

\section{Iterative Training with Tailored Samples}
In this section, we first present our concept of samples in the context of minimal path method, and describe the differences between samples and annotations.
After that, we show details of our schemes for inference and training.

\subsection{Sample and Annotation}
\begin{figure}[!t]
  \centering
  \begin{tabular}{ccc}
    \includegraphics[width=2.2cm]{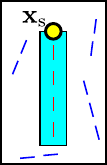}&
    \includegraphics[width=2.2cm]{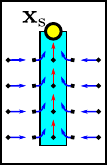}&
    \includegraphics[width=2.2cm]{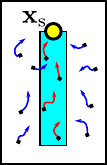}\\
    (a) Unrealistic  & (b) Semi-realistic & (c) Realistic
  \end{tabular}
  \caption{
    Sampling schemes.
    Positive and negative samples are shown as red and blue line segments.
    (a) In \pathcnn \cite{Liao_CVPR_2022_Progressive}, samples are drawn along lines (annotations). Such samples are unrealistic.
    (b) In the initial iteration of our method, samples are drawn using the predecessor function \dijPrev computed using the ground truth mask. They are not very realistic, but better than samples in (a).
    (c) In the main iterations of our method, samples are drawn using \dijPrev, which is computed using the CNN classifier trained in the previous iteration.
    Samples drawn in this way are more realistic.
  }
  \label{pic:sampling}
\end{figure}

An important reason for the inexact boundary in the segmentation mask in \cref{pic:pathcnn_boundary_problem} is the discrepancy between samples used for training and samples encountered actually during inference.
Similar to the description in \cref{sec:pathcnn}, a \emph{sample} in our method is an image patch defined by its centerline and a specific width.
Instead of using annotations such as segmentation masks or lines directly, we propose to \emph{derive} samples with much more semantic information.
Otherwise, the samples may be generated inappropriately.
For example, in the sampling scheme of \pathcnn, both positive and negative samples (image patches) are drawn along annotations (lines), i.e., annotations are used \emph{directly}.
The centerlines of such positive and negative samples are shown as red and blue line segments in \cref{pic:sampling}a, respectively.
The positive training samples lie on the centerlines, but during the minimal path computation (i.e., inference), real samples which should be classified as tubular structure are rarely exactly on the centerlines.
To deal with this discrepancy, we introduce an \emph{iterative} training scheme to generate training samples which are specifically \emph{tailored} for minimal path methods, without changing existing annotations.
Compared with the training samples in \cite{Liao_CVPR_2022_Progressive}, our training samples contain more semantic information to improve the minimal path method, since such training samples are likely to be more similar to samples encountered during inference (\cref{pic:sampling}b and c).
In the following, we describe the inference method first, because it is not only used during the inference phase, but also used during the training phase.

\subsection{Inference}
\label{sec:inference}

\begin{figure}[!t]
  \centering
  \includegraphics[width=8.5cm]{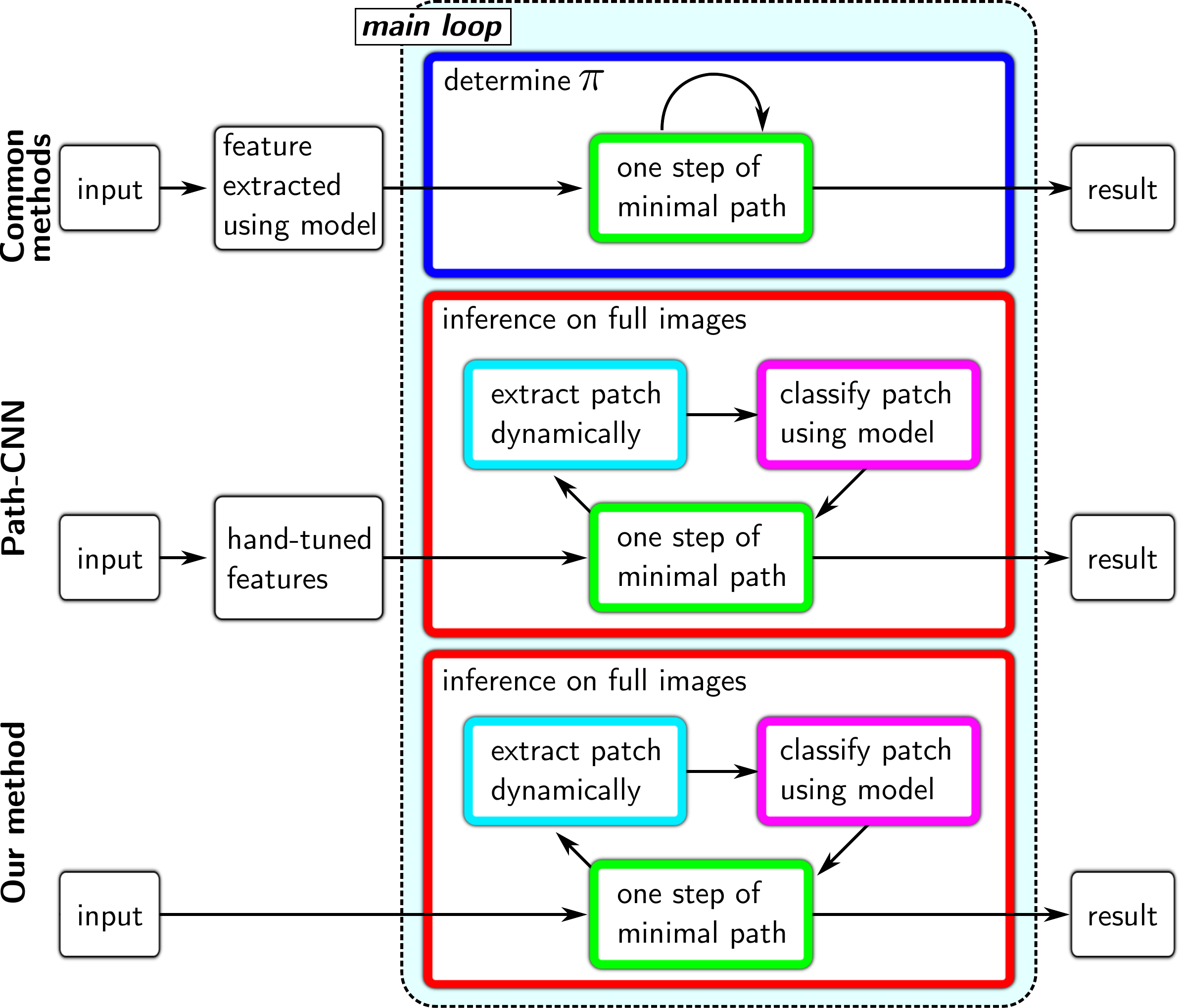}
  \caption{
    Comparison of the inference step in different minimal path methods.
    The operations (boxes with colored border) correspond to functions in \cref{alg:dijkstra} with the same name.
  }
  \label{pic:inference_scheme}
\end{figure}

The main steps of our inference method and a comparison with previous methods is given in \cref{pic:inference_scheme}.
In common minimal path methods, features are computed only once before the main loop of Dijkstra's algorithm.
In \pathcnn, hand-tuned features are used to initialize the graph.
But during the main loop, learning-based features are extracted using a CNN classifier to update the graph.
Our method, which is similar to \pathcnn but does not use hand-tuned features, is described as \cref{alg:dijkstra}.

\begin{algorithm}[!t]
  \LinesNumbered
  \SetKwData{result}{result}
  \SetKw{init}{Initialization}
  \KwIn{Image $\image$, start point $\dijStart$, CNN classifier \myclassifier}
  \KwOut{predecessor function $\dijPrev$, segmentation mask \mymask}

  \lFor{each $e \in \dijEdge$}{
    $\dijW[e] \gets \epsilon$\label{func:init_weight}
  }
  \For{each $\position\in\dijVertex$}{ \label{func:init_start}
    $\dijPrev[\position] \gets \dijNone$;
    $\dijDist[\position] \gets \infty$;
    $\mymask[\position] \gets \dijFG$;
  }
  $\dijDist[\dijStart] \gets 0$; $\dijQ \assign \dijVertex$; $\dijS \assign \emptyset$\label{func:init_done}\;

  \While(\tcp*[h]{Main loop}){$\dijQ \not= \emptyset$}{\label{func:main_loop_start}
    $u \assign \argmin_{\position\in\dijQ}\dijDist(\position)$\label{func:select_u}\;
    $\dijQ \gets \dijQ - \{u\}$; $\dijS \assign \dijS \cup \{u\}$\label{func:move_u}\;
    $P \gets \dijGetRectifiedPatch(u, \dijPrev, \image)$\label{func:get_patch}\;
    $\dijClass \gets \dijClassify(\myclassifier, P)$\label{func:classify}\;
    $\mymask[u] = \dijClass$\label{func:assign_label}\;
    \For{each $v \in \dijN(u) \cap \dijQ$}{

      \If{$\dijClass = \dijBG$}{\label{func:need_dynamic}\
        $\dijW[e_{u,v}] \gets \dijW[e_{u,v}] + \dijPenalty$\label{func:increase_weight}\;
      }
      \If{$\dijDist[v] > \dijDist[u] + \dijW[e_{u,v}]$}{\label{func:update_d1}
        $\dijDist[v] \assign \dijDist[u] + \dijW[e_{u,v}]$\label{func:update_d2}\;
        $\dijPrev[v] \assign u$\label{func:update_d3}\;
      }
    }
  }\label{func:main_loop_end}
  \Return $\dijPrev, \mymask$
  \caption{\ApplyClassifier: Minimal path method with dynamic weight adaptation.
    %
  }
  \label{alg:dijkstra}
\end{algorithm}
The input of our algorithm includes the input image, a start point, and a CNN classifier.
We initialize all edge weights with a small constant $\epsilon$ (\cref{func:init_weight}).
The functions \dijPrev and $d$, an empty segmentation mask \mymask, as well as regions \dijS and \dijQ are initialized between \cref{func:init_start} and \ref{func:init_done}.
\cref{func:main_loop_start} to \ref{func:main_loop_end} are the main loop of the inference, which only stops if \dijPrev is determined for all pixels in the image, i.e., if \dijQ is empty.
In each iteration, the pixel $u$ with currently minimum $d$ value is moved from \dijQ to \dijS (\cref{func:select_u} and \ref{func:move_u}), and a patch \mypatch starting at $u$ is extracted \emph{dynamically} (\cref{func:get_patch}), i.e., this patch can only be extracted during the main loop but not before it.
Patch extraction is also illustrated in \cref{pic:dijkstra}a.
The patch \mypatch is then \emph{rectified}, i.e., transformed into a rectangular patch so that its centerline becomes straight and vertical, and classified by the CNN classifier (\cref{func:classify}).
The classification label is added into the mask \mymask (\cref{func:assign_label}).
In case of background label, the weights between $u$ and its neighbors inside \dijN (the 4-neighborhood on the image grid) and \dijQ are increased by a high penalty \dijPenalty (\cref{func:increase_weight}), so that it is less possible for the minimal path to run through $u$.
After that, functions \dijPrev and $d$ are updated as in the usual version of Dijkstra's algorithm (\cref{func:update_d1} to \ref{func:update_d3}).
Finally, \dijPrev and \mymask are returned.
\mymask is the segmentation mask of \emph{all} the tubular structures in the image, and \dijPrev can be used to extract the minimal path between start point \fmstartpoint and \emph{any} other point in the image, especially the end point \fmendpoint.
The centerlines extracted using \dijPrev have line-topology, while \mymask does not have this property.

\subsection{Training}
\begin{figure}[!t]
  \centering
  \includegraphics[width=8.2cm]{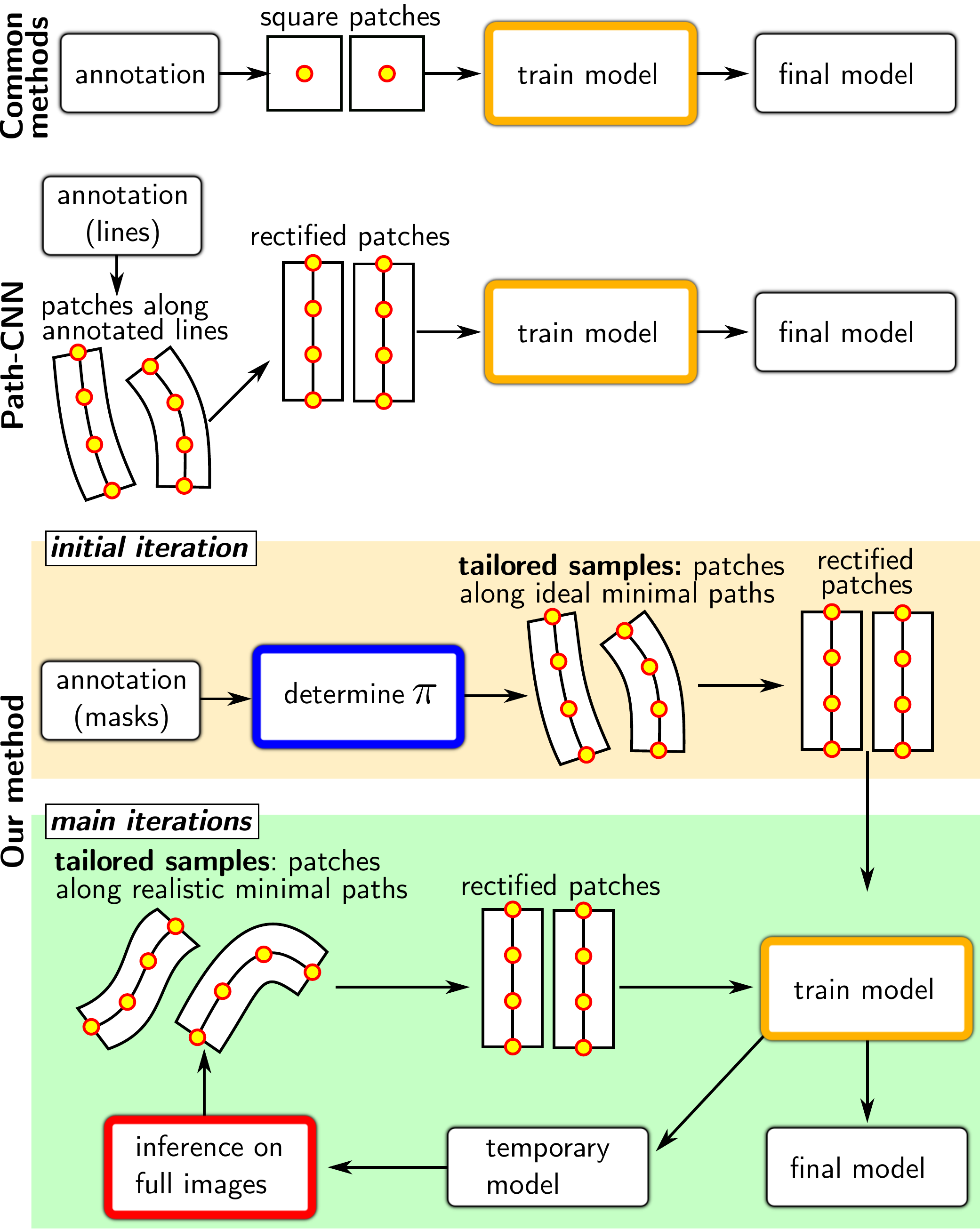}
  \caption{
    Comparison of the training step in different minimal path methods.
    Each yellow circle with red border corresponds to a pixel.
    The operations (boxes with colored border) correspond to functions in \cref{alg:training} with the same name.
    The red and blue boxes are the same as the red and blue boxes in \cref{pic:inference_scheme}.
  }
  \label{pic:training_scheme}
\end{figure}

To use more realistic samples for training, we apply an iterative scheme.
The main steps and comparison with other training methods are shown in \cref{pic:training_scheme}.
In particular, while common methods and \pathcnn draw samples directly from the annotations, we use minimal path method itself to get initial samples and then repeatedly refine the samples.
In the $N$-th iteration, the minimal path method with trained classifier $\myclassifier_{N-1}$ from the $(N-1)$-th iteration is applied to the training images to extract samples.
Details of our method is given in \cref{alg:training}.

For simplicity reasons, we assume there is only one training image and one validation image, each with ground truth mask.
\begin{algorithm}[!tb]
  \LinesNumbered
  \KwIn{Train/val images \trainingimage and \validationimage with ground truth \traininggt, \validationgt, start points \fmstartpointtrain, \fmstartpointval, maximum number of iterations $N$}
  \KwOut{Classifier \myclassifier}

  \tcp*[h]{Initial iteration}

  $\dijPrev \gets \DeterminePi(\traininggt, \fmstartpointtrain)$\label{func:dummy_weight}\;
  $\mysamples \gets \CreateTailoredSamples(\dijPrev, \trainingimage, \traininggt)$\label{func:dummy_tailor}\;
  $\diceprev \gets 0$\;

  \For(\tcp*[h]{Main iterations}){$i\gets 1$ \KwTo $N$}{

    $\myclassifier \gets \TrainClassifier(\mysamples)$\label{func:classifier}; \label{func:train_classifier}

    $\dijPrev, \mymask \gets \ApplyClassifier(\trainingimage, \fmstartpointtrain, \myclassifier)$\label{func:weight}\;

    $\mysamples \gets \CreateTailoredSamples(\dijPrev, \trainingimage, \traininggt)$\label{func:tailor}\;

    \tcp{Validation}
    $\dijPrev, \mymask \gets \ApplyClassifier(\validationimage, \fmstartpointval, \myclassifier)$\label{func:apply_classifier}\;
    $\dicecurrent \gets \ComputeDice(\mymask, \validationgt)$\label{func:compute_dice}\;
    \eIf {$\dicecurrent > \diceprev$}{\label{func:validation_start}
      $\diceprev \gets \dicecurrent$;
    }{\textbf{break}}\label{func:validation_end}

  }
  \Return \myclassifier
  \caption{Iterative training scheme
  }
  \label{alg:training}
\end{algorithm}
In contrast to \pathcnn, hand-tuned features are not used in our method.
Instead, in \DeterminePi (\cref{func:dummy_weight}) we use \emph{ground truth} mask to define initial weights for edges in the graph: If both vertices of an edge belong to foreground, then weight of the edge is set to 1.
Otherwise, it is set to a large positive constant.
This graph is used to obtain \dijPrev \emph{without} using any classifier.
We then use \dijPrev to extract \emph{tailored samples} from the \emph{training image} \trainingimage (\cref{func:dummy_tailor}): Starting from any pixel \position, we generate a path \mypathlocal by back-tracing using \dijPrev for a fixed number of steps, and crop a curved image patch \mypatch along \mypathlocal.
The sample \mypatch generated in this way is positive if \position in in foreground.
Otherwise, it is negative.
The path \mypathlocal is a segment on the minimal path between \position and \fmstartpointtrain.
\mypathlocal is computed not using weights based on image features, but using weights derived from ground truth, so its shape is still quite regular (\cref{pic:sampling}b), but it is more realistic than samples extracted using manually specified lines (\cref{pic:sampling}a).
With these samples, an initial classifier is trained (\cref{func:train_classifier}).
The steps in \cref{func:weight} and \ref{func:tailor} is very similar to the steps in \cref{func:dummy_weight} and \ref{func:dummy_tailor}, but now \dijPrev is computed using training image and classifier, instead of using ground truth.
The samples generated in this way are getting even closer to the realistic samples, since both of them are generated by applying the same inference method, i.e., minimal path method with integrated CNN.
These samples are illustrated in \cref{pic:sampling}c and \cref{pic:training_scheme}.
The trained classifier is applied for inference on the validation image (\cref{func:apply_classifier}) to compute a segmentation mask \mymask.
The iteration terminates if the Dice score between \mymask and ground truth does not increase anymore (\cref{func:compute_dice} to \ref{func:validation_end}).

\section{Experimental Results}
We applied our method to three datasets and compared the results with seven previous methods.
The accuracy of centerlines and segmentation masks is studied below, along with qualitative results.

\subsection{Datasets}
\label{sec:datasets}
Three public datasets \dataepfl  \cite{Sironi_PAMI_2015}, \datatoronto \cite{MnihThesis}, and \datadrive \cite{Staal2004} were used for evaluation.
In most of our experiments, the training sets consisted of quite small numbers of images.
\paragraph{\dataepfl} This dataset contains 14 satellite images of roads.
Binary segmentation masks and centerlines for all roads are provided.
The centerlines are presented as sequences of image coordinates.
We used 3 images for training and the remaining 11 images for testing.
\vspace*{-1pt}
\paragraph{\datatoronto} This dataset consists of 1157 satellite images of roads, divided into a training set of 1108 images, and a test set of 49 images.
Only the road centerlines are provided, but they are presented as binary masks instead of sequences of coordinates.
We dilated the centerlines slightly and used the dilated region as an approximation of the foreground mask.
Regions close to the foreground mask were not used for training.
The regions further away from the foreground were used as background mask.
We used 10 image patches for training and apply the trained model to the test set.
\vspace*{-1pt}
\paragraph{\datadrive} In this dataset, there are 40 images of retinal vessels.
The dataset is divided into 20 training images and 20 test images.
Binary masks for the vessels are provided without centerlines.

\subsection{Baselines}
Beside minimal path methods, U-Net and its variants are also widely used to segment tubular structures.
Thus, in addition to minimal path methods, i.e., \dcnn \cite{Dijkstra59:NM}, \pcnn  \cite{Liao_PAMI_2018_Final}, and \pathcnn \cite{Liao_CVPR_2022_Progressive}, we also compared our method with the original \abbrunet \cite{Ronneberger_MICCAI_2015_U-Net}, \abbrunet with softDice loss \cite{Milletari_3DV_2016} and clDice loss \cite{Shit_CVPR_2021_clDice}, and \frunet \cite{Liu_JBHI_2022_FR-UNet_RetinalVessel} (one of the best methods to segment retinal vessels \cite{paper_with_code_retina}).
In each experiment, all methods were trained using the same images.

\dcnn is the classical Dijkstra's algorithm, and edge weights in the graph are determined using a CNN before Dijkstra's algorithm is applied.
\pcnn is the progressive minimal path method, which also uses a CNN to pre-compute edge weights.
In contrast to \dcnn, \pcnn extracts a short \emph{path segment} $\gamma$ in every iteration and adapts the edge weights on-the-fly, depending on the CNN-based features along  $\gamma$.
\pathcnn is an extension of \pcnn: In each iteration, not a path segment but an \emph{image patch} is extracted (\cref{pic:dijkstra}a) to adapt edge weights.
The inference step of our method is similar to \pathcnn (\cref{pic:inference_scheme}), although the training step is very different (\cref{pic:training_scheme}).
The CNN used in \dcnn, \pcnn, \pathcnn and our method was MobileNetV2 \cite{Sandler_CVPR_2018_MobileNetV2}.
For \pathcnn and our method, we used a fixed patch size of $31\times31$ for all experiments and datasets.
For \dcnn and \pcnn, the usual data augmentation operations were used, while for \pathcnn and our method we used only horizontal and vertical flips, and a small rotation of $\pm 5$ degrees, since further augmentations like rotation or random cropping would violate the assumption of these methods that the centerline of a rectified patch must be a \emph{straight vertical line} in the middle of the patch (\cref{sec:pathcnn}).
We compared these minimal path methods in terms of the accuracy of the centerline points.
To evaluate the accuracy of the segmentation masks, our method was compared with \pathcnn, \abbrunet, softDice, clDice, and \frunet.

\subsection{Accuracy of the Centerlines}
We used the mean error between result and ground truth to evaluate methods which extract lines as sequences of points.
This error measure is defined as
\begin{equation}
\errormeasure = \frac{1}{N}\sum_{\mypath}\sum _{\position_i\in \mypath} |\position_i - \refpoint{i}|,
\label{eqn:error_measure}
\end{equation}
where $N$ is the total number of control points on the extracted centerlines, \mypath is a centerline, and \refpoint{i} is the ground truth point which is closest to a given point $\position_i$ on \mypath.

To compare the approaches, we selected 100 paths out of 11 test images of \dataepfl, and 650 paths out of 49 test images of \datatoronto.
The results are summarized in \cref{table:centerline}.
\begin{table}[!t]
  \small
  \begin{center}
    \begin{tabular}{c|c|c|c|g}
      \hline
      \textbf{Dataset} & \textbf{\dcnn} & \textbf{\pcnn}  & \textbf{\pathcnn}      & \textbf{Ours}\\\hline\hline
      \dataepfl        &  2.91          & 2.52            & 1.71                   & \textbf{1.41}\\\hline
      \datatoronto     &  7.36          & 5.12            & 3.65                   & \textbf{3.54}\\\hline
    \end{tabular}
  \end{center}
  \vspace*{-5pt}
  \caption{Errors for centerlines measured using \cref{eqn:error_measure}.  %
    Unit: Pixel.
  }
  \label{table:centerline}
\vspace{-8pt}
\end{table}
All four methods used features obtained using the same CNN architecture, but the performance was rather different.
In \dcnn, the CNN is applied to each pixel individually, and the relationship between neighboring pixels is only taken into account by convolutional filters.
Since tubular structures are typically thin and long, they are difficult to capture using convolutional filters alone, leading to high errors.
\pcnn computes features explicitly for path segments and yielded mean errors of $2.52$ and $5.12$ pixels for \dataepfl and \datatoronto, respectively, while in both datasets, the width of most roads was over 10 pixels.
By using features of rectified patches, \pathcnn and our method were able to incorporate more contextual information, and further reduced the errors to less than 2 pixels for \dataepfl and less than 4 pixels for \datatoronto.
Overall, our method achieved the lowest errors.

\begin{figure}[!t]
  \centering
  \begin{tabular}{cccc}
    \includegraphics[width=3.8cm, angle=180]{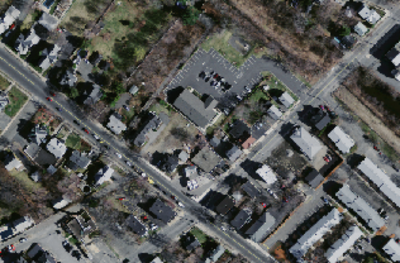}&
    \includegraphics[width=3.8cm, angle=180]{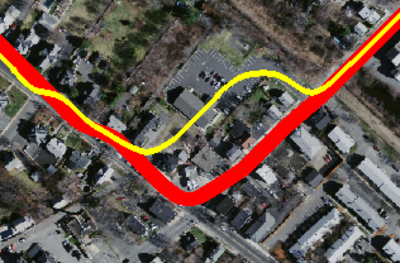}\\
    &\\
    \includegraphics[width=3.8cm]{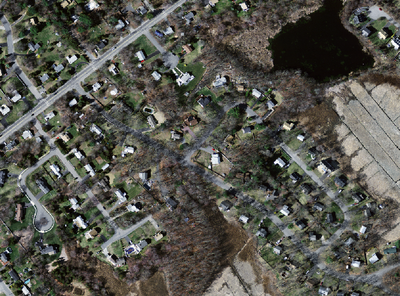}&
    \includegraphics[width=3.8cm]{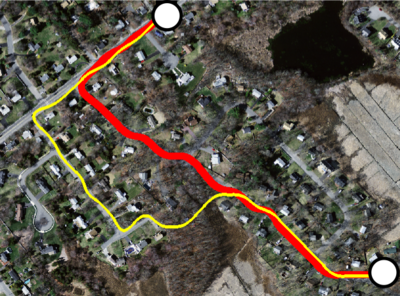}\\
    (a) Input & (b) Results
  \end{tabular}
  \caption{
    Determination of the path between given start and end points on two images from \datatoronto.
    Result of \pathcnn is shown as yellow line, and our result is shown as red line.
    \pathcnn made several wrong turns, while our method found the correct paths.
  }
  \label{pic:compare_centerlines}
\end{figure}

By using samples tailored for minimal path methods, our method performed better in complex environments.
This is demonstrated in the qualitative comparison between \pathcnn and our approach in \cref{pic:compare_centerlines}.
In the upper row, \pathcnn made a wrong turn at a crossing, while in the lower row it failed to follow the correct road which had strong shadow.
In contrast, our method dealt well with both cases.

\subsection{Accuracy of the Segmentation Masks}
{Dice score} is a commonly used metric to compare two masks $\mymask_1$ and $\mymask_2$.
It is defined as
\begin{equation}
\dicescore = 2 \cdot \frac{|\mymask_1  \cap \mymask_2|}{|\mymask_1|  + |\mymask_2|},
\label{eqn:dice_score}
\end{equation}
where $|\mymask_1  \cap \mymask_2|$ is the number of pixels in the overlapping region.
For \dataepfl and \datadrive, we computed the mean Dice scores (\meandice) of the test sets, i.e., the Dice scores are computed for each test image individually and then averaged.

\begin{figure}[!t]
	\centering
	\begin{tabular}{cccc}
		\includegraphics[width=3.8cm]{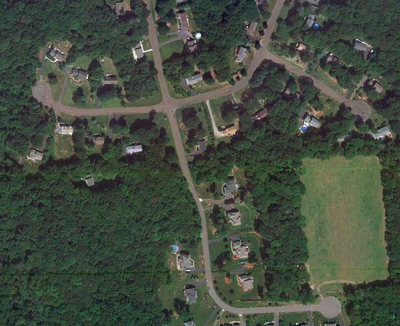}&
		\includegraphics[width=3.8cm]{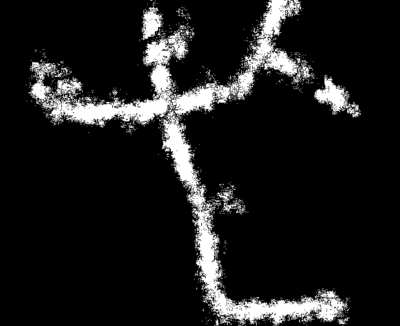}\\
		(a) Input & (b) \pathcnn \\
		\includegraphics[width=3.8cm]{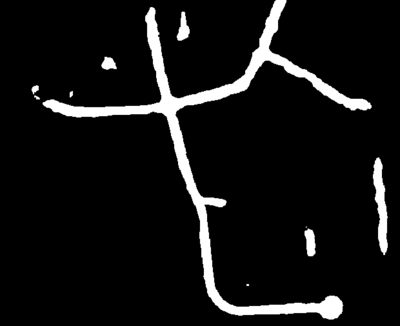}&
		\includegraphics[width=3.8cm]{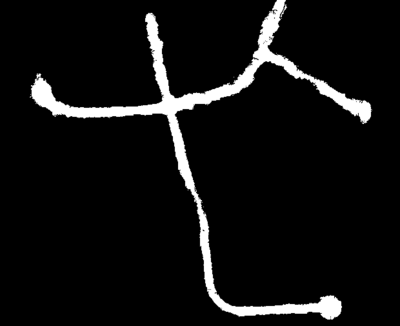}\\
		(c) \frunet & (d) Our method
	\end{tabular}
	\caption{
		Results for a test image from \dataepfl.
		All three models are trained with only 3 images.
		\pathcnn could not determine the boundary precisely, and \frunet found false positive regions.
	}
	\label{pic:compare_epfl}
	\vspace*{-4pt}
\end{figure}

\begin{figure}[!t]
  \centering
  \begin{tabular}{cc}
    \includegraphics[width=3.8cm]{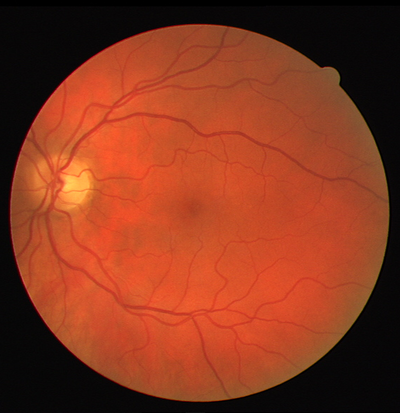}&
    \includegraphics[width=3.8cm]{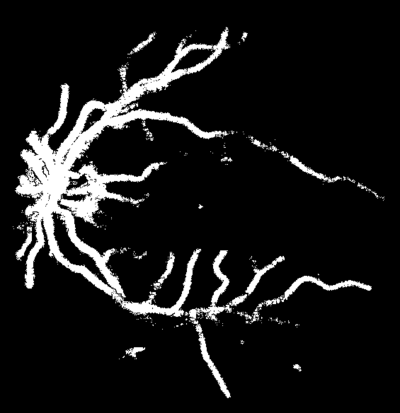}\\
    (a) Input image & (b) \pathcnn\\
    \includegraphics[width=3.8cm]{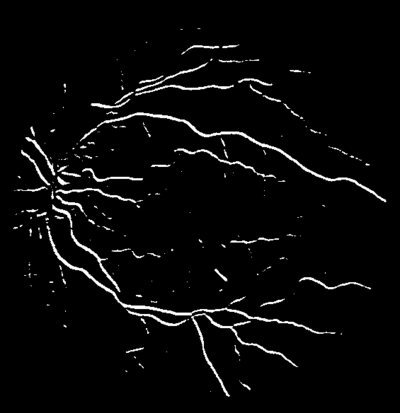}&
    \includegraphics[width=3.8cm]{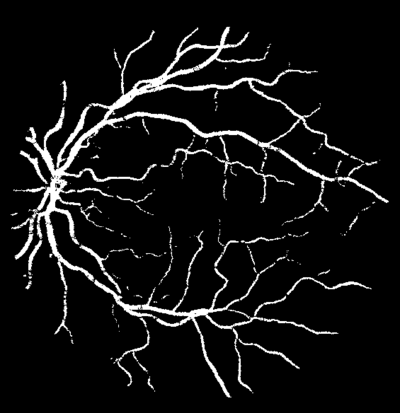}\\
    (c)  \abbrunet & (d) Our method
  \end{tabular}
  \caption{
    Results for a test image from \datadrive.
    All three models were trained with only 3 images.
    \pathcnn and \abbrunet both failed to segment vessels in image regions with low contrast, while our method successfully segmented most vessels.
  }
  \label{pic:compare_mask_drive}
  \vspace{-5pt}
\end{figure}

The results are summarized \cref{table:binary_mask}.
\begin{table}[!t]
  \small
  \begin{center}
    \setlength\tabcolsep{3pt}
    \begin{tabular}{c|c|c|c|c|c|g}
      \hline
      \scriptsize\textbf{Dataset} & \scriptsize\textbf{\pathcnn}    &\scriptsize\textbf{\abbrunet} & \scriptsize\textbf{\softdice} &\scriptsize\textbf{\cldice} & \scriptsize \textbf{\frunet}  & \scriptsize\textbf{Ours} \\\hline\hline
      \dataepfl        & 0.6322               & 0.9076            & 0.8803         & 0.9052         & 0.8847            & \textbf{0.9125}     \\\hline
      \datadrive       & 0.5927               & 0.7815            & 0.7644         &  0.7888        & \textbf{0.8208}            & 0.8065     \\\hline
    \end{tabular}
  \end{center}
  \vspace*{-5pt}
  \caption{Dice score for segmentation measured using \cref{eqn:dice_score}.
  }
  \label{table:binary_mask}
\end{table}
For \dataepfl, we achieved the highest \meandice.
Also, the boundary of tubular structures was delineated more precisely using our method, as illustrated in \cref{pic:compare_epfl}.
The masks obtained using \pathcnn have quite noisy boundary, while in the result of \frunet, several background regions are segmented as foreground.
In contrast, our method successfully avoided false positive regions, and achieved smoother boundary than \pathcnn.
For \datadrive, our \meandice was slightly below \frunet but higher than all other methods.
In addition, we ran a further experiment using only 3 training images and all test images on \datadrive.
Our \meandice was still as high as $0.7578$, while the \meandice of \abbrunet and \pathcnn dropped to $0.5217$ and $0.5470$, respectively.
A comparison is shown in \cref{pic:compare_mask_drive}.
Our method not only achieved more precise boundary, but also made the segmentation of tubular structures more robust in regions with low image quality, especially when the training data was scarce.
In the example in \cref{pic:compare_mask_drive}a, a large part on the vessel tree is missing in the results of \pathcnn and \abbrunet, especially in the right half of the image, where the image contrast is relatively low.
In contrast, our method correctly segmented most vessels, even the small ones.
These results were achieved using only 3 training images, which is much less than the amount of annotations needed by most recent methods.
It shows that our method is especially interesting for medical image analysis, where it is often difficult to obtain enough annotated images.

\subsection{Effects of Iterative Training}

\begin{figure}[!t]
  \centering
   \includegraphics[width=8cm]{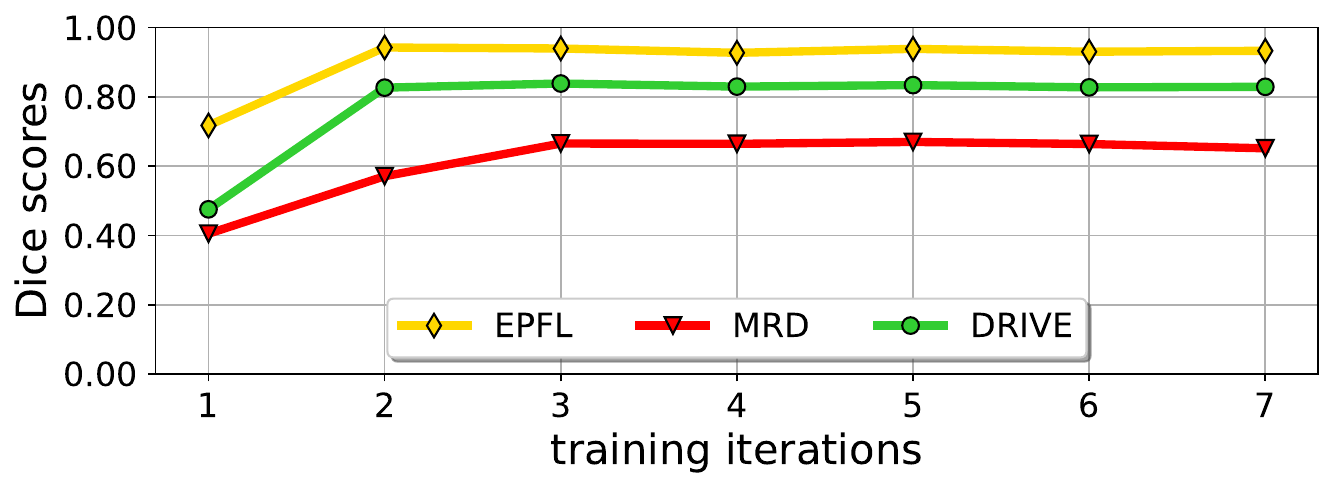}
  \caption{
   Dice scores on the validation set after each iteration.
  }
  \label{pic:dice-change}
\vspace{-7pt}
\end{figure}
The Dice scores on the validation set after each training iteration are plotted in \cref{pic:dice-change}.
For all three datasets, the Dice score increases fast in the first iterations, and after three iterations there is no significant change anymore.
Note that there are no ground truth masks for \datatoronto, so we used dilated centerlines as pseudo-masks to train our method (\cref{sec:datasets}).
Therefore \datatoronto has lower Dice score than the other datasets in \cref{pic:dice-change}.
\begin{figure}[!t]
  \centering
  \begin{tabular}{ccc}
    \includegraphics[width=2.4cm]{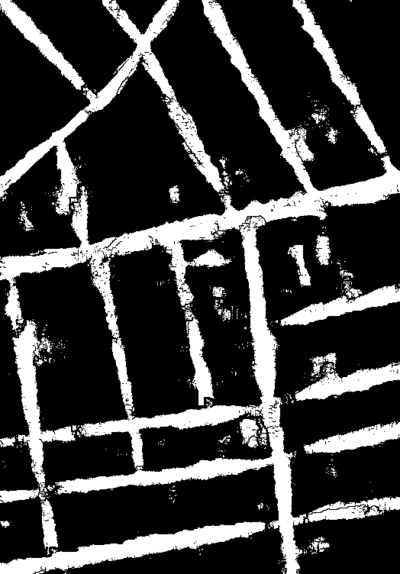}&
    \includegraphics[width=2.4cm]{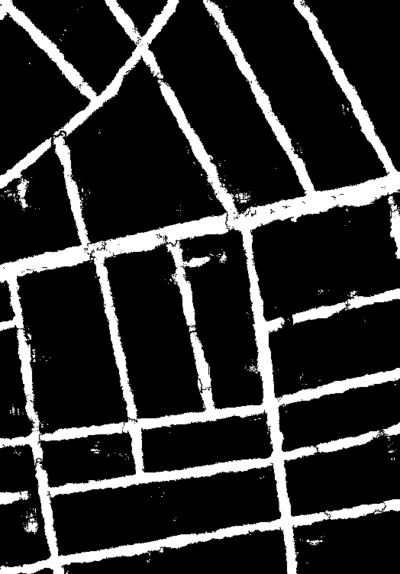}&
    \includegraphics[width=2.4cm]{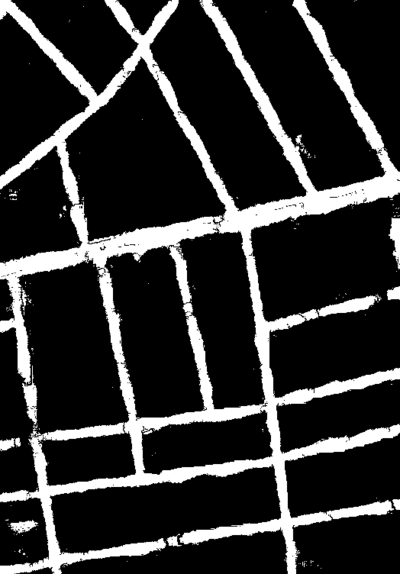}\\
    (a) & (b) & (c) \\
    \includegraphics[width=2.4cm]{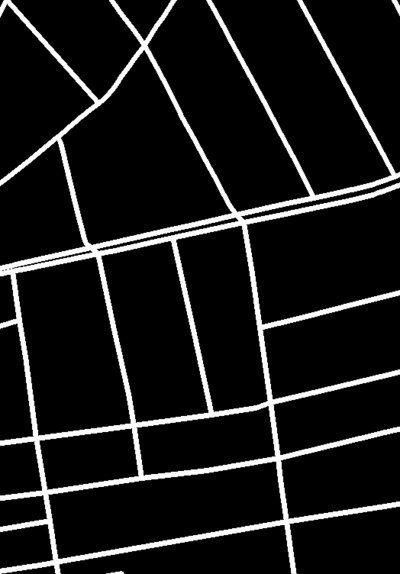}&
    \includegraphics[width=2.4cm]{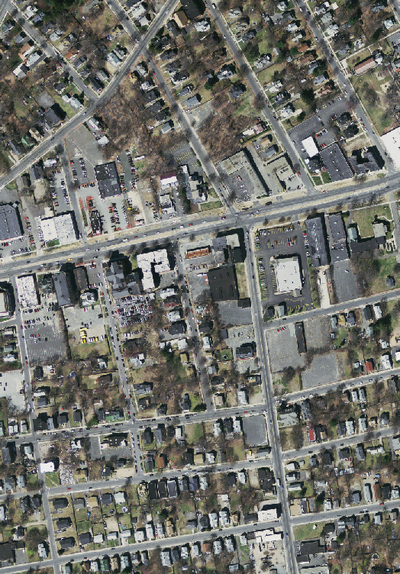}&
    \includegraphics[width=2.4cm]{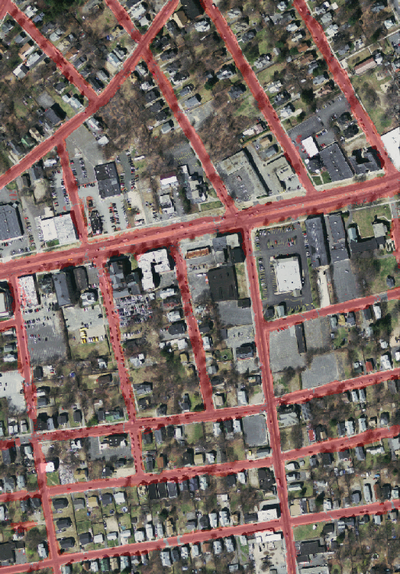}\\
    (d) & (e) & (f)
  \end{tabular}
  \caption{
    Improvement through the iterative training scheme for \datatoronto.
    Segmentation mask after the first, second, and third iteration are shown in (a), (b), and (c) respectively. (d) Ground truth (centerline only). (e) Input image. (f) Overlay of the result after the third iteration on the input image.
  }
  \label{pic:patches}
\end{figure}
A qualitative examples in \cref{pic:patches} shows the improvement of segmentation masks due to the iterative training scheme, using a test image in \datatoronto .
After the first iteration, the boundary of the roads is still quite noisy, and there are several false positives in the background.
Also, the width of the roads is imprecise and fluctuates a lot.
After the third iteration, false positives are reduced significantly, and an overlay with the original image shows that the shape of the roads is determined well, even though our method is trained only using pseudo-masks.

\subsection{Implementation and Run Time} The major part of our C++ implementation runs on CPU, only the step of patch classification runs on GPU.
It takes about 15 minutes to segment an image of the size $600\times 600$ using a 3.4 GHz Intel i7 CPU and NVIDIA GTX 1070 GPU.

\section{Conclusion}
We proposed a minimal path method to segment tubular structures and extract their centerlines.
An \emph{iterative training scheme} is introduced so that existing annotations can be used more effectively without modification.
In our framework, minimal path method is not only used for segmentation and centerline extraction, but also for the generation of realistic training samples.
Our minimal path method contains an integrated CNN classifier.
Annotations are not directly used as samples to train this classifier.
Instead, the minimal path method is used along with the classifier from the previous iteration to extract training samples out of existing annotations.
Trained with these samples, the classifier in turn is used to steer the minimal path method in the next iteration to generate better samples.
Training samples obtained in this way are semantically more meaningful and better \emph{tailored} for the minimal path method, so our model is trained more effectively.
Experimental results show that by using this training scheme, our approach outperforms several recent methods and yields high-quality centerlines and segmentation masks for tubular structures on several datasets in different domains, even in cases of very limited amounts of annotations.
We believe that our method is especially useful for application areas in which annotations are difficult to obtain, such as medical image processing.

{\small
\bibliographystyle{ieee_fullname}
\bibliography{literature}

\begin{thebibliography}{10}\itemsep=-1pt

\bibitem{paper_with_code_retina}
{Leaderboard of Retinal Vessel Segmentation on DRIVE}.
\newblock
  \url{https://paperswithcode.com/sota/retinal-vessel-segmentation-on-drive},
  2023.
\newblock [Online; accessed 08. March 2023].

\bibitem{Bastani_CVPR_2018}
Favyen Bastani, Songtao He, Sofiane Abbar, Mohammad Alizadeh, Hari
  Balakrishnan, Sanjay Chawla, Sam Madden, and David DeWitt.
\newblock {RoadTracer: Automatic Extraction of Road Networks from Aerial
  Images}.
\newblock In {\em IEEE Conference on Computer Vision and Pattern Recognition},
  2018.

\bibitem{Batra_CVPR_2019_RoadJointLearning}
Anil Batra, Suriya Singh, Guan Pang, Saikat Basu, C.V. Jawahar, and Manohar
  Paluri.
\newblock Improved road connectivity by joint learning of orientation and
  segmentation.
\newblock In {\em IEEE Conference on Computer Vision and Pattern Recognition},
  2019.

\bibitem{Benmansour2009b}
Fethallah Benmansour and Laurent~D. Cohen.
\newblock Fast object segmentation by growing minimal paths from a single point
  on 2d or 3d images.
\newblock {\em Journal of Mathematical Imaging and Vision}, 33(2):209 -- 221,
  2009.

\bibitem{Benmansour_IJCV_2011}
Fethallah Benmansour and Laurent~D. Cohen.
\newblock Tubular structure segmentation based on minimal path method and
  anisotropic enhancement.
\newblock {\em International Journal of Computer Vision}, 92:192--210, 2011.

\bibitem{Chen_TIP_2019}
Da Chen, Jiong Zhang, and Laurent~D. Cohen.
\newblock Minimal paths for tubular structure segmentation with coherence
  penalty and adaptive anisotropy.
\newblock {\em IEEE Transactions on Image Processing}, 28(3):1271--1284, Mar.
  2019.

\bibitem{Cohen_IJCV_1997}
Laurent~D. Cohen and Ron Kimmel.
\newblock Global minimum for active contour models: A minimal path approach.
\newblock {\em International Journal of Computer Vision}, 24(1):57--78, 1997.

\bibitem{Cormen2009}
Thomas~H. Cormen, Charles~E. Leiserson, Ronald~L. Rivest, and Clifford Stein.
\newblock {\em Introduction to Algorithms (Third Edition)}.
\newblock MIT Press, 2009.

\bibitem{Dijkstra59:NM}
E.~W. Dijkstra.
\newblock {A Note on Two Problems in Connection with Graphs}.
\newblock {\em Numerische Mathematik}, 1:269--271, 1959.

\bibitem{Frangi1998}
Alejandro~F. Frangi, Wiro~J. Niessen, Koen~L. Vincken, and Max~A. Viergever.
\newblock Multiscale vessel enhancement filtering.
\newblock In {\em Medical Image Computing and Computer-Assisted Intervention},
  1998.

\bibitem{Freiman2009}
M. Freiman, N. Broide, M. Natanzon, L. Weizman, E. Nammer, O. Shilon, J. Frank,
  L. Joskowicz, and J. Sosna.
\newblock Vessels-cut: A graph based approach to patient-specific carotid
  arteries modeling.
\newblock In {\em Proceedings of the Workshop on 3D Physiological Human}, 2009.

\bibitem{Hu_NIPS_2019_TopologyPreservingDeepImageSegmentation}
Xiaoling Hu, Fuxin Li, Dimitris Samaras, and Chao Chen.
\newblock Topology-preserving deep image segmentation.
\newblock In {\em Advances in Neural Information Processing Systems}, 2019.

\bibitem{Kaul-Yezzi-Tsai_PAMI_2012}
Vivek Kaul, Anthony Yezzi, and Yichang Tsai.
\newblock Detecting curves with unknown endpoints and arbitrary topology using
  minimal paths.
\newblock {\em IEEE Transactions on Pattern Analysis and Machine Intellegence},
  34(10):1952--1965, 2012.

\bibitem{Kleinberg2005}
Jon Kleinberg and Eva Tardos.
\newblock {\em Algorithm Design}.
\newblock Pearson Education, 2005.

\bibitem{Li2007}
Hua Li and Anthony Yezzi.
\newblock Vessels as 4-d curves: Global minimal 4-d paths to extract 3-d
  tubular surfaces and centerlines.
\newblock {\em IEEE Transactions on Medical Imaging}, 26(9):1213--1223, 2007.

\bibitem{Liao_CVPR_2022_Progressive}
Wei Liao.
\newblock Progressive minimal path method with embedded cnn.
\newblock In {\em IEEE Conference on Computer Vision and Pattern Recognition},
  2022.

\bibitem{Liao_PAMI_2018_Final}
Wei Liao, Stefan Wörz, Chang-Ki Kang, Zang-Hee Cho, and Karl Rohr.
\newblock Progressive minimal path method for segmentation of 2d and 3d line
  structures.
\newblock {\em IEEE Transactions on Pattern Analysis and Machine Intelligence},
  40(3):696--709, Mar. 2018.

\bibitem{Liu_JBHI_2022_FR-UNet_RetinalVessel}
Wentao Liu, Huihua Yang, Tong Tian, Zhiwei Cao, Xipeng Pan, Weijin Xu, Yang
  Jin, and Feng Gao.
\newblock {Full-Resolution Network and Dual-Threshold Iteration for Retinal
  Vessel and Coronary Angiograph Segmentation}.
\newblock {\em IEEE Journal of Biomedical and Health Informatics},
  26(9):4623--4634, 2022.

\bibitem{Milletari_3DV_2016}
Fausto Milletari, Nassir Navab, and Seyed-Ahmad Ahmadi.
\newblock V-net: Fully convolutional neural networks for volumetric medical
  image segmentation.
\newblock In {\em International Conference on 3D Vision}, 2016.

\bibitem{MnihThesis}
Volodymyr Mnih.
\newblock {\em {Machine Learning for Aerial Image Labeling}}.
\newblock PhD thesis, University of Toronto, 2013.

\bibitem{Mosinska_PAMI_2020}
Agata Mosinska, Mateusz Kozinski, and Pascal Fua.
\newblock Joint segmentation and path classification of curvilinear structures.
\newblock {\em IEEE Transactions on Pattern Analysis and Machine Intellegence},
  2020.

\bibitem{Pechaud2009}
Mickael Pechaud, Renaud Keriven, and Gabriel Peyre.
\newblock Extraction of tubular structures over an orientation domain.
\newblock In {\em IEEE Conference on Computer Vision and Pattern Recognition},
  2009.

\bibitem{Ronneberger_MICCAI_2015_U-Net}
Olaf Ronneberger, Philipp Fischer, and Thomas Brox.
\newblock {U-Net: Convolutional Networks for Biomedical Image Segmentation}.
\newblock In {\em Medical Image Computing and Computer-Assisted Intervention},
  2015.

\bibitem{Sandler_CVPR_2018_MobileNetV2}
Mark Sandler, Andrew Howard, Menglong Zhu, Andrey Zhmoginov, and Liang-Chieh
  Chen.
\newblock Mobilenetv2: Inverted residuals and linear bottlenecks.
\newblock In {\em IEEE Conference on Computer Vision and Pattern Recognition},
  2018.

\bibitem{Sato1998}
Yoshinobu Sato, Shin Nakajima, Nobuyuki Shiraga, Hideki Atsumi, Shigeyuki
  Yoshida, Thomas Koller, Guido Gerig, and Ron Kikinis.
\newblock Three-dimensional multi-scale line filter for segmentation and
  visualization of curvilinear structures in medical images.
\newblock {\em Medical Image Analysis}, 2(2):143--168, 1998.

\bibitem{Sethian1999}
J.~A. Sethian.
\newblock {\em Level Set Methods and Fast Marching Methods: Evolving Interfaces
  in Computational Geometry, Fluid Mechanics, Computer Vision, and Materials
  Science}.
\newblock Cambridge University Press, 1999.

\bibitem{Shit_CVPR_2021_clDice}
Suprosanna Shit, Johannes~C. Paetzold, Anjany Sekuboyina, Ivan Ezhov, Alexander
  Unger, Andrey Zhylka, Josien P.~W. Pluim, Ulrich Bauer, and Bjoern~H. Menze.
\newblock cldice - a novel topology-preserving loss function for tubular
  structure segmentation.
\newblock In {\em IEEE Conference on Computer Vision and Pattern Recognition},
  2021.

\bibitem{Sironi_PAMI_2015b}
Amos Sironi, Bugra Tekin, Roberto Rigamonti, Vincent Lepetit, and Pascal Fua.
\newblock Learning separable filters.
\newblock {\em IEEE Transactions on Pattern Analysis and Machine Intelligence},
  37(1):94--106, 2015.

\bibitem{Sironi_PAMI_2015}
Amos Sironi, Engin Tueretken, Vincent Lepetit, and Pascal Fua.
\newblock Multiscale centerline detection.
\newblock {\em IEEE Transactions on Pattern Analysis and Machine Intellegence},
  2015.

\bibitem{Staal2004}
Joes Staal, Michael~D. Abramoff, Meindert Niemeijer, Max~A. Viergever, and Bram
  van Ginneken.
\newblock Ridge-based vessel segmentation in color images of the retina.
\newblock {\em IEEE Transactions on Medical Imaging}, 23(4):501--509, 2004.

\bibitem{Tueretken_ICCV_2013}
Engin Tueretken, Carlos Becker, Przemyslaw Glowacki, Fethallah Benmansour, and
  Pascal Fua.
\newblock Detecting irregular curvilinear structures in gray scale and color
  imagery using multi-directional oriented flux.
\newblock In {\em IEEE International Conference on Computer Vision}, 2013.

\bibitem{Ulen_PAMI_2015}
Johannes Ulen, Petter Strandmark, and Fredrik Kahl.
\newblock Shortest paths with higher-order regularization.
\newblock {\em IEEE Transactions on Pattern Analysis and Machine Intellegence},
  37(12):2588--2600, 2015.

\end{thebibliography}
}

\end{document}